\documentclass[letterpaper, 10 pt, conference]{ieeeconf}  

\IEEEoverridecommandlockouts                              

\usepackage{cite}
\usepackage{amsmath,amssymb,amsfonts}
\usepackage{algorithmic}
\usepackage{graphicx}
\usepackage{textcomp}
\usepackage{xcolor}
\usepackage{multirow}
\usepackage{url}
\usepackage{booktabs}
\def\BibTeX{{\rm B\kern-.05em{\sc i\kern-.025em b}\kern-.08em
    T\kern-.1667em\lower.7ex\hbox{E}\kern-.125emX}}

\usepackage{acronym}
\usepackage{balance}

\title{\LARGE \bf
A Case Study on the Acceptance of a Humanoid Robotic Head Employed in Three
Public Spaces
}

\author{Marcel Heisler$^{1}$, Luca Randecker$^{1}$, and Christian Becker-Asano$^{1}$
\thanks{$^{1}$All authors are with Stuttgart Media University, Germany \newline {\tt\small heisler|randecker|becker-asano}\newline {\tt\small @hdm-stuttgart.de}
}}

\usepackage[colorinlistoftodos]{todonotes}
\acrodef{TAM2}{Technology Acceptance Model Version 2}
\acrodef{ML}{Machine Learning}
\acrodef{STT}{Speech-to-Text}
\acrodef{TTS}{Text-to-Speech}
\acrodef{ASR}{Automatic Speech Recognition}
\acrodef{RAG}{Retrieval Augmented Generation}
\acrodef{HRI}{Human-Robot Interaction}
\acrodef{ITU}{intention to use}
\acrodef{PU}{perceived usefulness}
\acrodef{PEOU}{perceived ease of use}

\begin{document}

\bstctlcite{IEEEexample:BSTcontrol}

\maketitle
\thispagestyle{empty}
\pagestyle{empty}

\begin{abstract}
Previous research has shown that a human-like robot's acceptance heavily depends on the setting in which it operates and its ability to perform relevant tasks.

This paper, first, reports on how our robot processes natural language to generate a multimodal, verbal response integrating emotional expressions based on an emotion simulation backend. Then, it describes how visitors were invited to speak with our robot in their own language at three different, public locations, where the robot was running continuously for several days. The \textit{\ac{TAM2}} questionnaire results reveal that on average users were motivated to use the robot and found it rather useful and easy to use regardless of the specific location. However, public spaces like the tourist information and the city library seem to be a better fit for our interactive, robotic head than an office environment such as the building authority, where the willingness to interact was lower. Overall, the robot's multi-lingual responses were very much appreciated, but every fifth user found the response time too slow impeding the dialog flow, which remains to be improved in future work.
\end{abstract}

\section{INTRODUCTION}

Whether human-like robots will become a part of everyday life in the future depends on several factors, especially their acceptance by humans \cite{graaf2013, broadbent2009} in different roles \cite{mavridis2012, hoshikawa2015}.
This, in turn, can be influenced by their appearance, their usefulness in the respective setting, their ability to perform the required tasks, as well as the expectations and attitudes of the people who interact with these robots \cite{goetz2003, broadbent2009, broadbent2013, graaf2013, mara2015, haring2016, wirtz2018, hall2019, naneva2020, song2022}. Research also highlights that cultural norms significantly shape their acceptance \cite{li2010, graaf2013}.

Among other factors, human-like robots are accepted by society if they are considered useful in the respective setting \cite{graaf2013, song2022, hall2019, naneva2020}. Several use cases with human-like robots have already been investigated, especially with people in need of a specific service \cite{mende2019}.

To further investigate how they are perceived by people in different public spaces, our android robot head, named Kim, spent three weeks interacting with visitors at the \textit{tourist information} (five days), the \textit{building authority} (four days), and the \textit{city library} (five days) in Stuttgart, Germany.
Some visitors provided data to evaluate the acceptance, in particular the usefulness of the robot at each location based on a German translation of the \acf{TAM2} \cite{venkatesh2000, olbrecht2010}.

Therefore, the remainder of this paper is structured as follows: Section~\ref{sec:related_work} outlines previous work related to human-like robots employed in public spaces, Section~\ref{sec:methods} describes the methods used to gather the data, Section~\ref{sec:results} presents the results followed by Section~\ref{sec:limitations} discussing potential limitations. Section~\ref{sec:discussion_and_conclusion} summarizes the findings and highlights opportunities for future research.

\section{Related Work}\label{sec:related_work}
Humanoid robots are increasingly being deployed in public spaces such as airports, hotels, and touristic places to enhance customer service and explore new forms of \ac{HRI} \cite{tong2024}.
While \textit{humanoid} robots represent a broader category with a wide range of applications, this work focuses specifically on \textit{android} robots, a subclass of humanoid robots designed to closely resemble humans \cite{ishiguro2006}.
Hence, there are fewer studies with android robots than studies involving humanoid robots. However, in public spaces, android robots have been employed in a variety of roles, such as salespeople in clothing stores \cite{watanabe2015}, receptionists in public and private institutions \cite{watanabe2014, kondo2011, liu2012, vishwanath2019, hashimoto2007, umetani2019, kobayashi2003, reis2020}, with \cite{umetani2019} specifically examining a library setting, HR experts \cite{stock2019, stock2020}, interviewers in recruitment contexts \cite{inoue2020, baka2022, inoue2021}, restaurant servants \cite{lu2021}; customer service agents in insurance companies \cite{mishra2019} information kiosks in museums \cite{heisler2025conversationsandreavisitorsopinions}, and in general service roles \cite{chuah2021}.

Nevertheless, none of these studies have conducted a comparative analysis with an android robot across different public places. Furthermore, no previous study has used a validated questionnaire based on the \acf{TAM2} \cite{venkatesh_theoretical_2000, olbrecht2010} to measure and compare the acceptance of an android robot in several real-world settings.

\section{Methods}\label{sec:methods}
\begin{figure*}[ht] 
\vspace{0.3cm}
\centering
    \centering
    \includegraphics[width=0.8\linewidth]{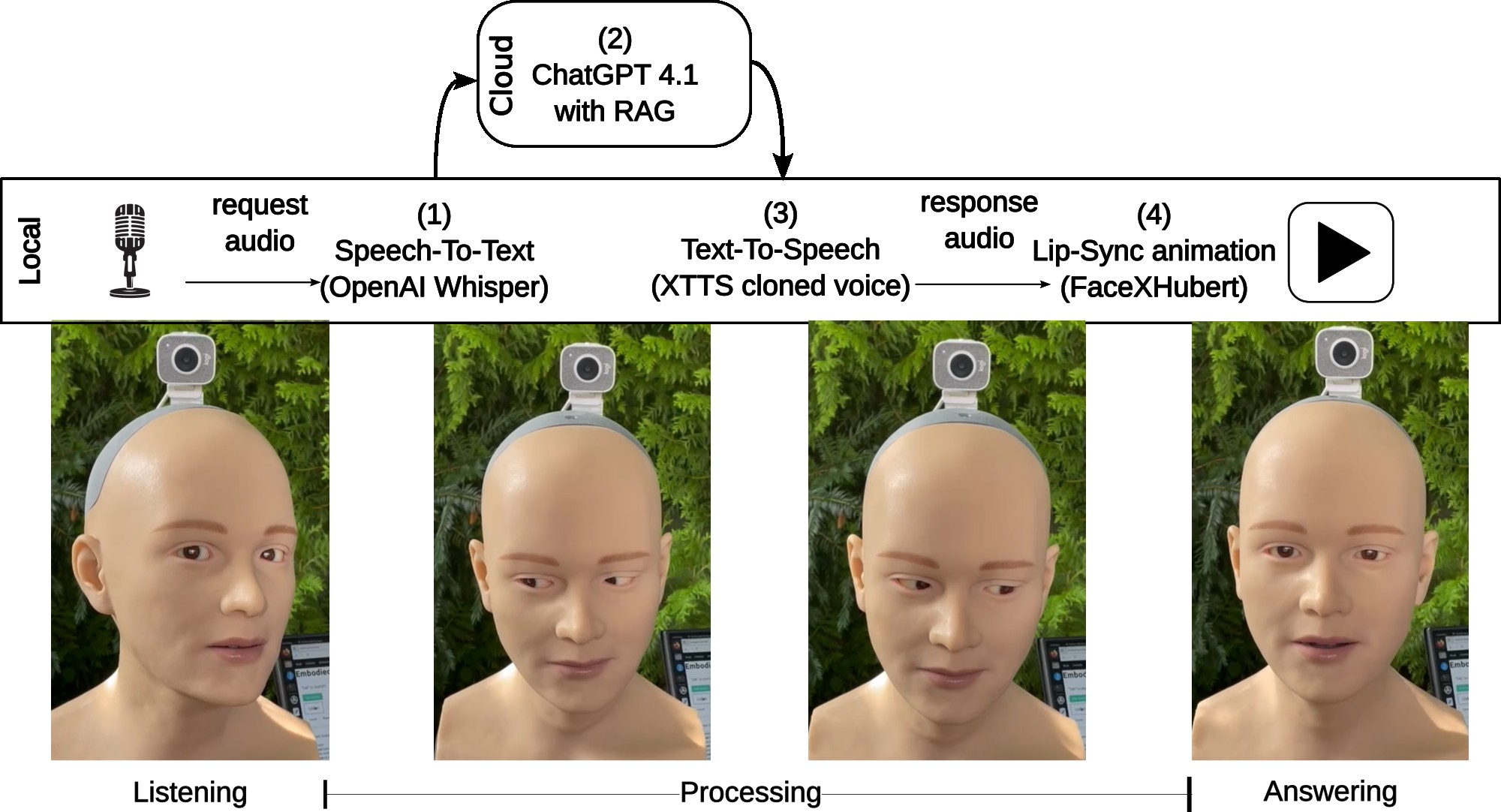}
    \caption{While the microphone in front of the robot is turned on, the robot head Kim performs a ``listening'' animation. Then, OpenAI's Whisper speech-to-text model is used to transcribe ``request audio'' into text. The resulting text is sent to an OpenAI ChatGPT 4.1 assistant, which uses Retrieval Augmented Generation (RAG) together with a location-specific system prompt to generate a textual answer. The text-to-speech software XTTS transforms the text into a ``response audio'', which is used by FaceXHubert to encode a lip-sync animation of the robot's face. Finally, the animation is played back together with the audio. The video stream provided by the camera behind the robot's head is analyzed for the presence of human faces, of which the closest is selected as an eye gaze target, whenever Kim is answering or waiting for user input.}
    \label{fig:softwarecomponents}
\end{figure*}

This section first describes the hardware and software components of the robot head, followed by the experimental setup at each location, the study procedure, and the measures that were applied.

\subsection{Hardware and Software}

As appearance and functionality are important factors for a human-like robot \cite{graaf2013, trieu2023}, we will provide information about Kim's hardware and software in detail below.

\subsubsection{Hardware}

The robot was produced by the same Japanese company that also produced the android robots \textit{Andrea} \cite{heisler2025conversationsandreavisitorsopinions}, \textit{Nikola} \cite{sato_android_2022}, \textit{Erica} \cite{glas_erica_2016} and \textit{Geminoid HI-1} \cite{von_der_putten_android_2011}. It features 14 pneumatic actuators that control the movements of its face, enabling expressive facial animations. 
To control its movements and expressions, 14 integer values between 0 and 255 are being sent via RS-485 with 25 hertz frequency by our custom developed software framework, see also \cite{heisler2023androidrobotheadembodied}. 


Unlike some of the aforementioned robots, Kim does not have cameras integrated into its eyes. Instead, a webcam is installed behind it to detect and look at the closest person. For audio interaction, an external microphone records sound input, while a loudspeaker provides output, ensuring basic conversational capabilities.

Our software framework runs on an \textit{Nvidia Jetson Orin}, while compressed air and electric power are used to drive the robot's actuators. The robot head is usually dressed in a black T-shirt and sometimes wears a baseball cap to hide the gray plate at the back of its head. 

\subsubsection{Software components} \label{sec:methods:software}

The software components are an extended version of \cite{heisler2025conversationsandreavisitorsopinions, heisler2023androidrobotheadembodied}, which combines four separate, open-source \ac{ML} models, see also Fig.~\ref{fig:softwarecomponents}:
\begin{enumerate}
    \item For \ac{STT} (step one) \verb|whisper-large-v3-turbo| is used supporting multiple languages as input.
    \item Multilingual output (step 3) is generated using XTTS \cite{casanova24_interspeech}  for speech synthesis. This \ac{ML} model supports 17 languages. A specific design-congruent voice \cite{kuch_your_2025} was cloned and used consistently in all languages.
    \item To speed up the generation of lip-sync animations (step 4) the former \ac{ML} model used as described in \cite{heisler_making_2023} was replaced by FaceXHubert \cite{haque_facexhubert_2023}.
    \item The people tracking is based on a realtime analysis of the video stream provided by the webcam using \verb|posenet| \cite{kendall_posenet_2015} (not shown in Fig.~\ref{fig:softwarecomponents}).
\end{enumerate}

To start a conversation users have to turn on the microphone, which also activates a listening animation. We opted against an active listening solution, because of a high probability of false activations due to the noisy environment in the public spaces. After the user turns off the microphone, the robot switches to a thinking animation, signaling that it is processing the input. As investigated in \cite{namba_how_2024}, thinking faces have a positive impact on a more natural \ac{HRI}. To further enhance its human-likeness the robot blinks at random and yawns if no interaction takes place for some time. 

To generate context-sensitive textual responses, three location-specific OpenAI assistants are used with their \verb|gpt-4.1| model as the basis. The assistants' prompts contain information about the robot itself and about its current location. In addition, \ac{RAG} is used to provide additional information (see Section~\ref{subsec:setup}). 

The robot's emotional states are simulated using WASABI \cite{becker-asano.2014}. To trigger the emotion dynamics of WASABI, the OpenAI assistant is prompted to calculate the valence of the last input sentence spoken by the user, which is then sent to WASABI as a valenced impulse ranging between -100 and +100. In effect, WASABI, which is running as a concurrent process in the same computer, returns the emotion likelihood of the emotions ``happy'', ``sad'', ``angry'', ``fearful'', ``disgusted'', ``surprised'', or ``neutral''. The internal emotion dynamic of WASABI lets the robot's emotional state automatically return to ``neutral'' after some time without any inputs. These emotions are expressed using validated static facial expressions \cite{kassner_comparing_2023}. The animations for thinking and speech have a higher priority than the emotional expressions for the relevant actuators. Thus, while the robot is speaking its mouth area is animated according to the speech signal, but before or after speech it is animated according to the emotion, e.g.~smiling.


\subsection{Experimental Setup} \label{subsec:setup}

\begin{figure*}[h] 
\centering
\begin{minipage}{0.32\textwidth}
    \centering
    \includegraphics[width=\linewidth]{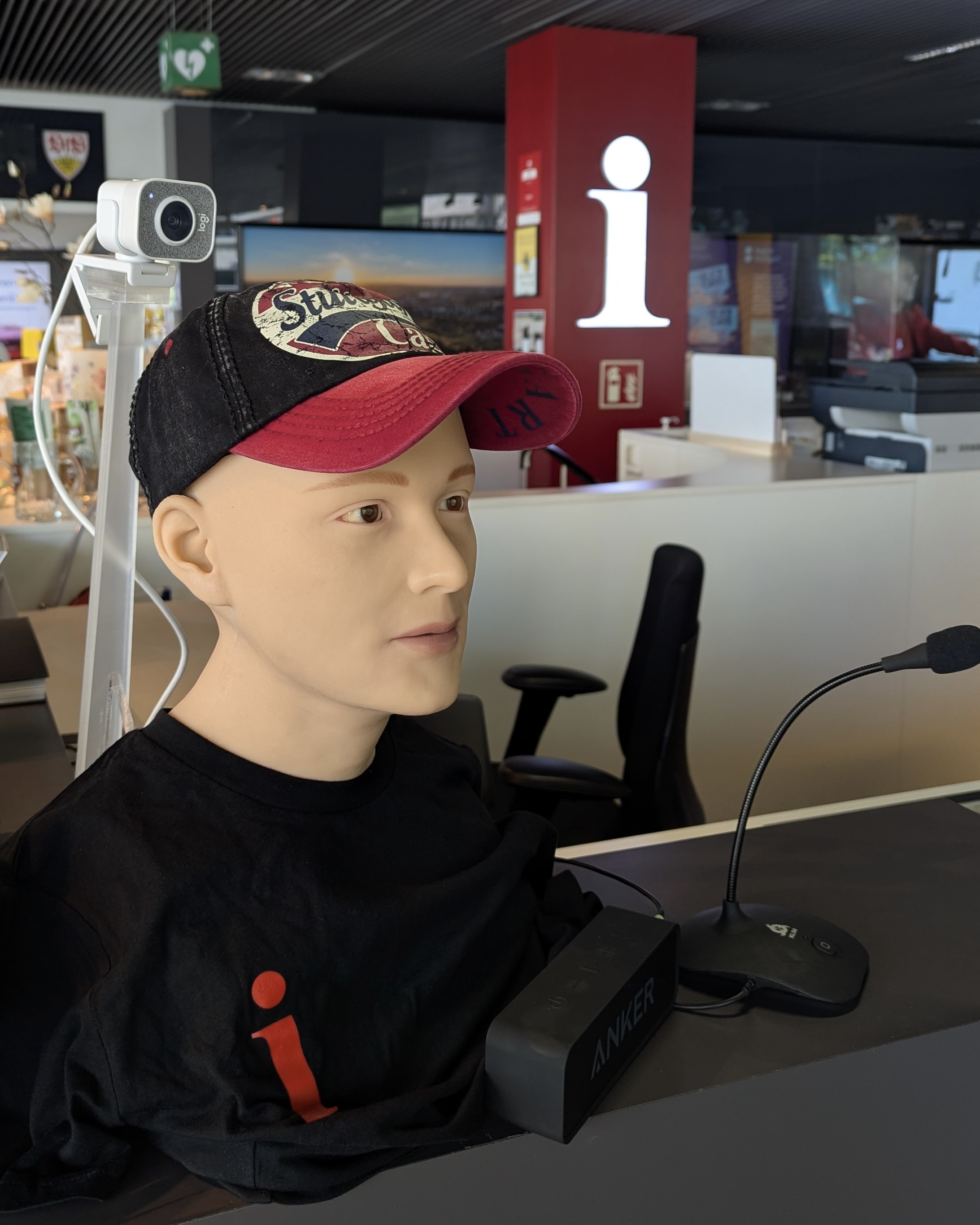}
    \caption{Robot head Kim at the tourist information in Stuttgart, Germany, available from May 26 to 30, 2025.} 
    \label{fig:touristinformation}
\end{minipage}\hfill
\begin{minipage}{0.32\textwidth}
    \centering
    \includegraphics[width=\linewidth]{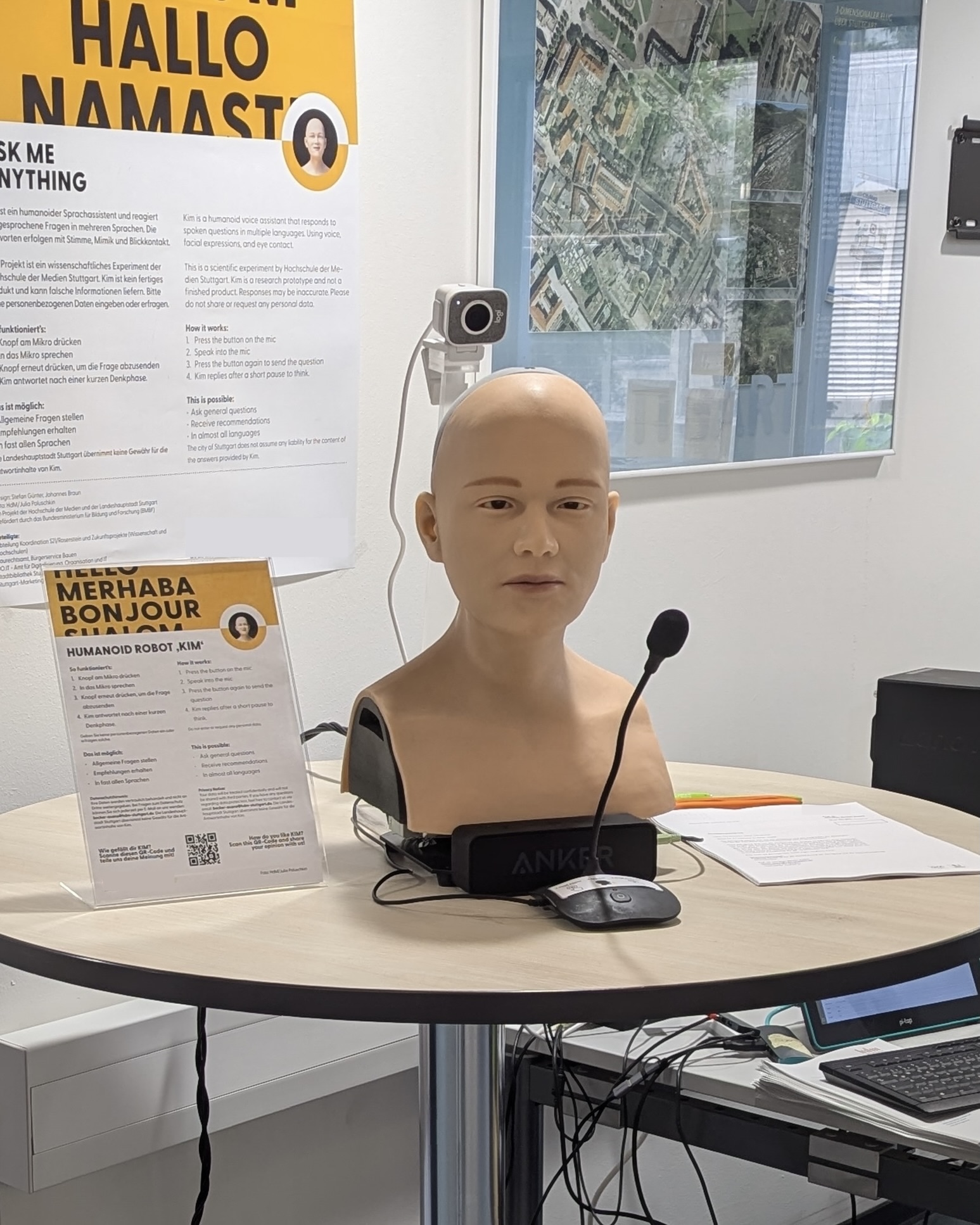}
    \caption{Robot head Kim at the building authority in Stuttgart, Germany, available from June 3 to 6, 2025.}
    \label{fig:buildingauthority}
\end{minipage}\hfill
\begin{minipage}{0.32\textwidth}
    \centering
    \includegraphics[width=\linewidth]{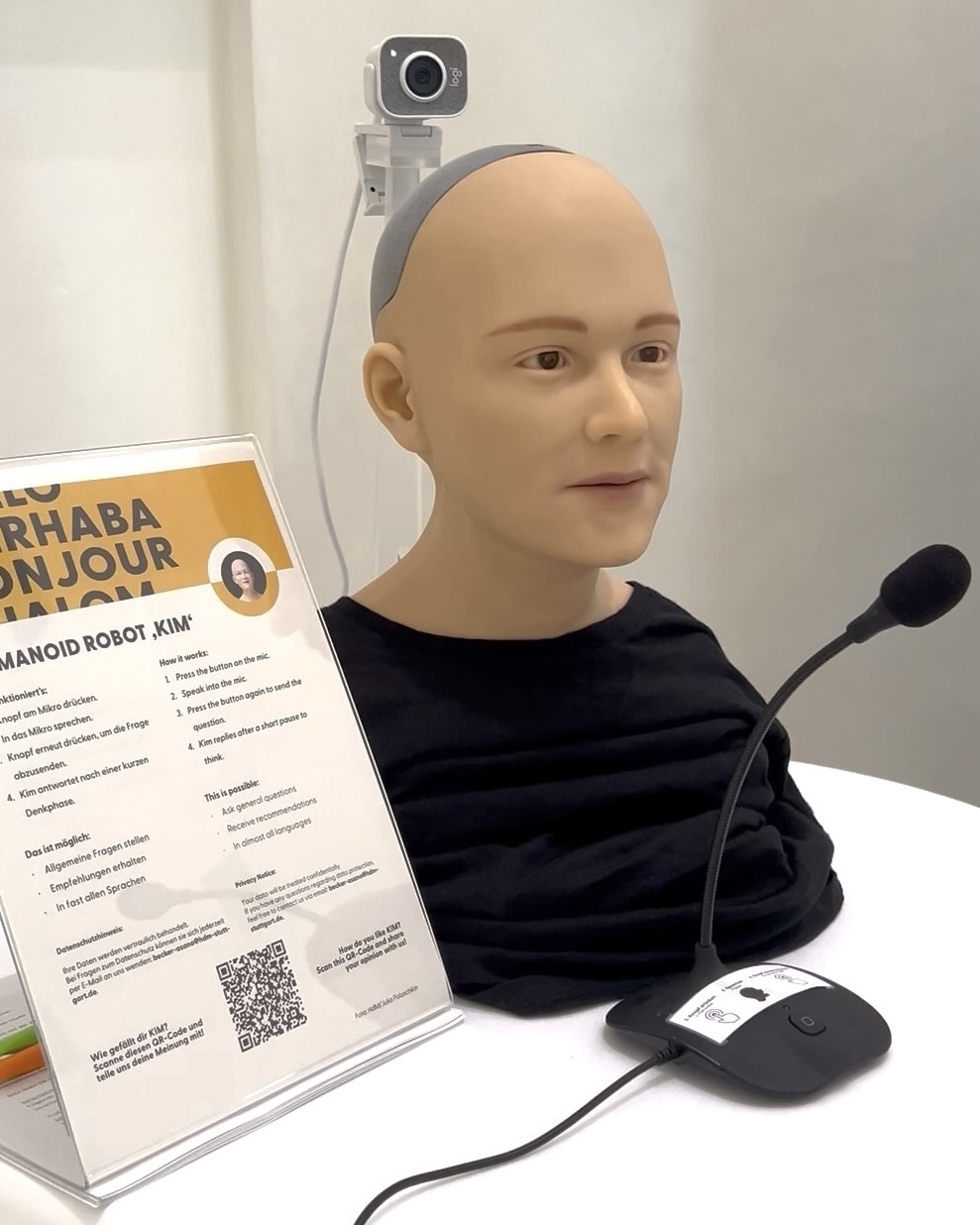}
    \caption{Robot head Kim at the city library in Stuttgart, Germany, available from June 10 to 14, 2025.}
    \label{fig:citylibrary}
\end{minipage}
\end{figure*}
Overall, the responses of the robot were shaped by the location-specific, publicly available information that was provided to the ChatGPT-based backend in advance. For the tourist information, the database contained information about the opening hours, tours, bus and train plans, tickets, services, and tips for visiting the city. At the building authority, the robot was provided with respective laws. The setting at the city library contained information about the opening hours, the building itself, and the locations of different book categories.

\subsubsection{Tourist Information}
The robot was present at the Stuttgart tourist information from May 26 to 30, 2025, between 10 AM and 6 PM each day. The tourist information is at a location with a lot of walk-in customers. A visible roll-up banner with information about Kim was placed directly to the left of the entrance. The robot head was placed on the front desk, as shown in Fig.~\ref{fig:touristinformation}, so that it was visible to visitors when they usually reached human staff to ask questions or make payments. Next to the head was a table display and underneath was a poster with information and instructions on how to interact with Kim.
During its deployment at this location, Kim wore a Stuttgart-branded cap and an employee t-shirt in black.

\subsubsection{Building Authority}
From June 3 to 6, 2025, Kim was at the building authority with the following schedule: Tuesday and Wednesday from 2 PM to 4 PM, Thursday from 9 AM to 6 PM, and Friday from 9 AM to 12 PM.
Unlike the other two locations, the building authority is not located in a busy area. In addition, the opening hours for visitors were heavily restricted, so that people that interacted with Kim were more likely to be employees in the same building than visitors with building concerns, which is also reflected in the number of completed questionnaires. However, there was also a roll-up, a table display and a poster inviting visitors to take notice of the robot. Kim itself was placed on a table approximately at the same height as the users, as shown in Fig.~\ref{fig:buildingauthority}.

\subsubsection{City Library}
From June 10 to 14, 2025, Kim was at the city library available to the public each day from 9 AM to 6 PM, located on the ground floor of the building, in an almost empty huge square space. Here, too, we made sure that the robot head was clearly visible and provided the relevant information, see Fig.~\ref{fig:citylibrary}. A lot of people have interacted with Kim, which is also reflected in the number of completed surveys.

\subsection{Study Procedure and Measurement}

\subsubsection{Study Procedure}
After the interactions users were invited to complete a questionnaire that contained an informed consent and items in both English and German. Additionally, participants were informed that they could cancel the survey at any time.

The number of participants at each location varies with 19 ($N_1=19$) completed questionnaires at the tourist information, seven ($N_2=7$) at the building authority, and 23 ($N_3=23$) at the city library. 

The \ac{TAM2} questionnaire from \cite{olbrecht2010} was used as a basis, as it provided the exact German translations of the original items \cite{venkatesh2000}. 
Our questionnaire can be divided into the following categories: general information such as location, date, time, and age; questions evaluating participants’ prior knowledge of the robot's availability at this place and whether it influenced their decision to visit the location; \ac{TAM2} items measuring  \textit{\ac{ITU}, \ac{PU}}, and \textit{\ac{PEOU}}; questions regarding the emotions; as well as additional questions, regarding for how long participants had interacted with the robot, which languages they used, and any suggestions they had. Finally, participants were asked how useful they considered the robot in this specific location. 

\subsubsection{Measurement}

Whether participants knew about the robot's availability prior to their visit, and whether they had specifically come to see the robot was evaluated with yes/no questions.

The \ac{TAM2} items and emotion-related questions were evaluated using a 7-point Likert scale, where 1 indicated `strongly disagree' and 7 indicated `strongly agree'. \ac{ITU} was measured with the items ``Assuming I have access to the robot, I intend to use it.'', and ``Given that I have access to the robot, I predict that I would use it.'' \ac{PU} was measured with four items ``Using the robot improves my performance.'', ``Using the robot increases my productivity.'', ``Using the robot enhances my effectiveness.'', and ``I find the robot to be useful.'' \ac{PEOU} was measured with four items ``My interaction with the robot is clear and understandable.'', ``My interaction with the robot does not require a lot of my mental effort.'', ``I find the robot to be easy to use.'', and ``I find it easy to get the robot to do what I want it to do.'' The \textit{mean value} for each category per participant was calculated. Additionally, the Shapiro-Wilk test was used to determine the $W$ statistic and $p$-value, in order to assess whether the data is normally distributed. 

The perceived emotionality of the robot head was measured with the items ``I have the impression that the robot displays emotional reactions.'', ``I find the robot’s emotional responses appropriate to the situation.'' and ``The robot appears moody or shows emotional mood swings.'' on the same 7-point Likert scale. The final question ``How useful do you think would it be to use this robot here today?'' was measured by a distinct Likert scale ranging from 0 `not at all' to 10 `very much' to compare our setup with a previous result of a study with the robot Andrea in a public museum \cite{heisler2025conversationsandreavisitorsopinions}.

\begin{figure*}[ht] 
\centering
\begin{minipage}{0.48\textwidth}
    \centering
    \includegraphics[width=0.7\linewidth]{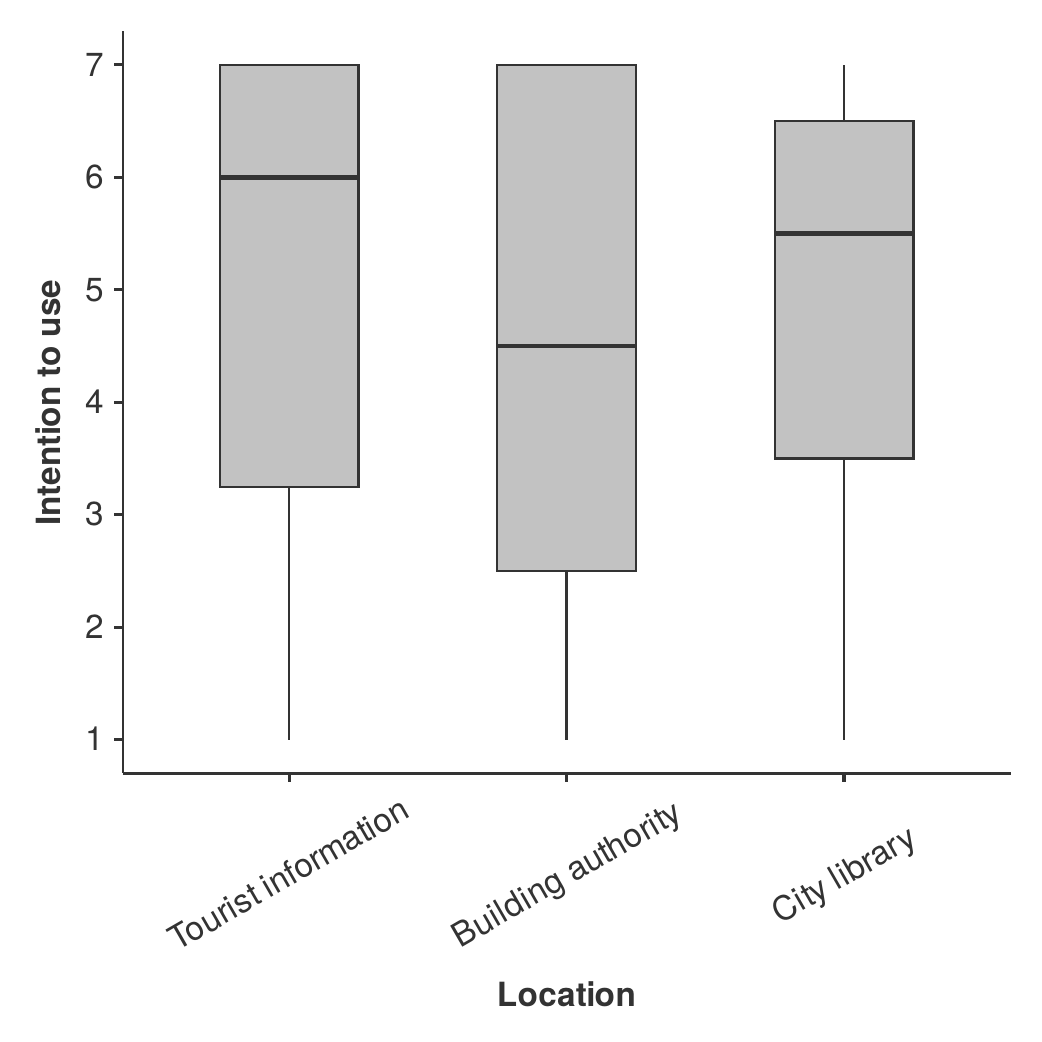}
    \caption{Box plot showing the distribution of \textit{\acf{ITU}} by location with indicated median values.}
    \label{intention_to_use_loc}
\end{minipage}\hfill
\begin{minipage}{0.48\textwidth}
    \centering
    \includegraphics[width=0.7\linewidth]{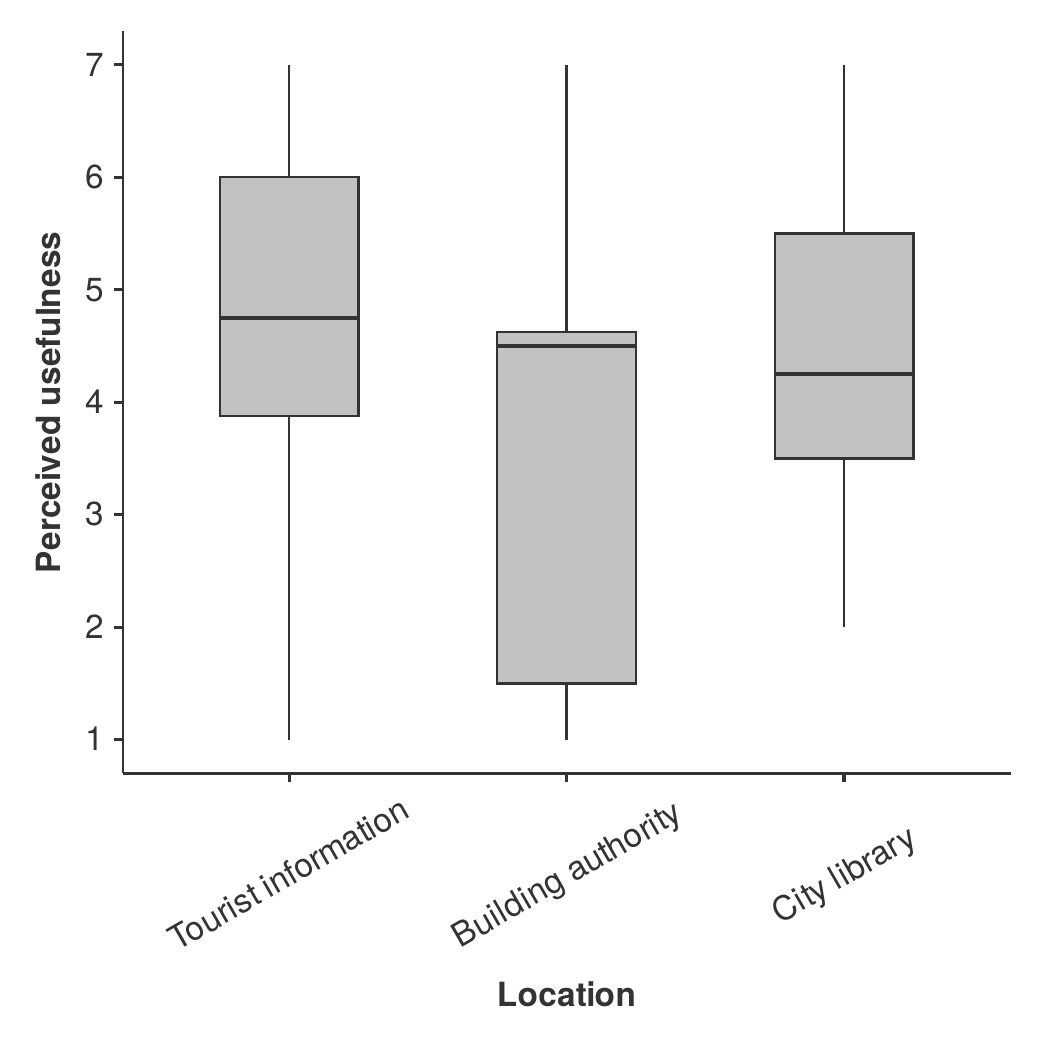}
    \caption{Box plot showing the distribution of \textit{\acf{PU}} by location with indicated median values.}
    \label{perceived_usefulness_loc}
\end{minipage}
\begin{minipage}{0.48\textwidth}
    \centering
    \includegraphics[width=0.7\linewidth]{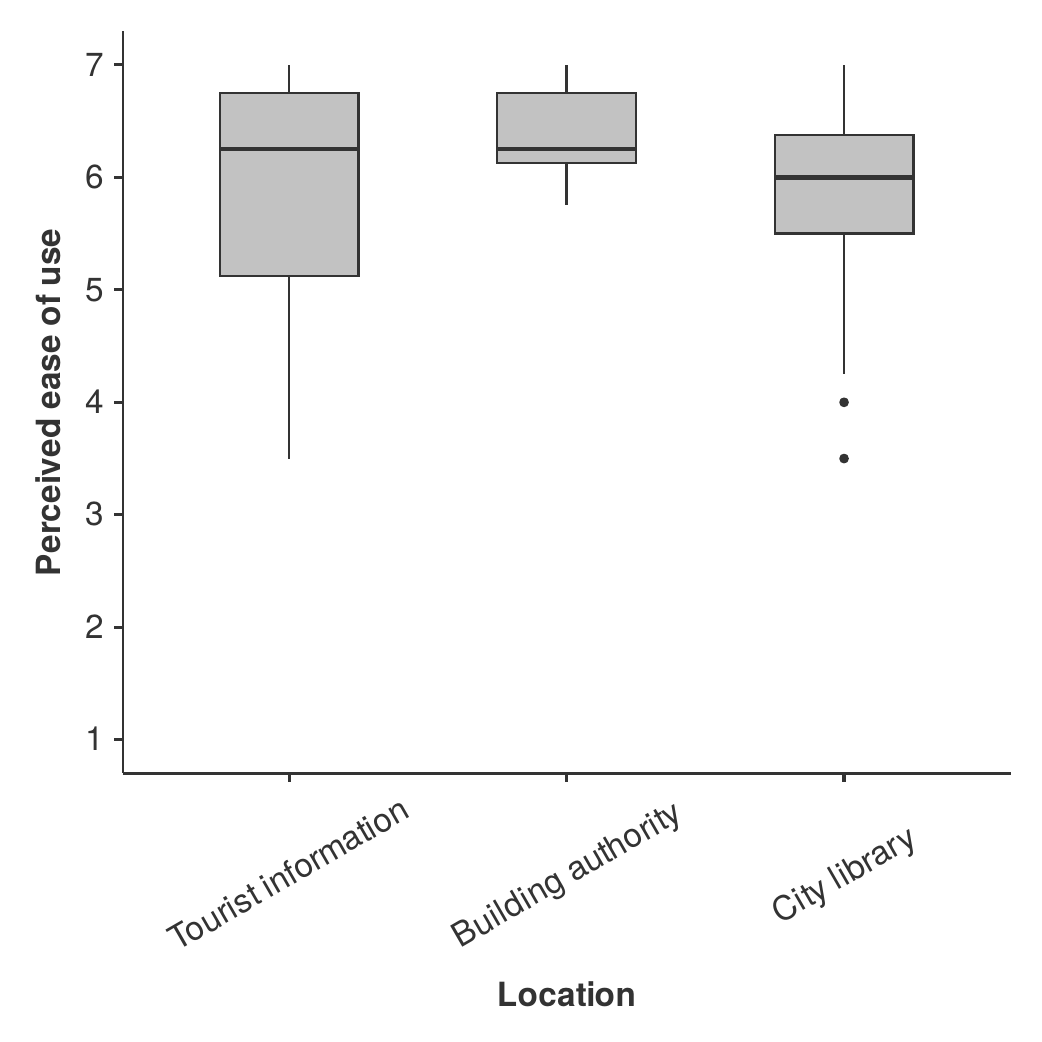}
    \caption{Box plot showing the distribution of \textit{\acf{PEOU}} by location with indicated median values.}
    \label{perceived_ease_of_use_loc}
\end{minipage}\hfill
\begin{minipage}{0.48\textwidth}
    \centering
    \includegraphics[width=0.7\linewidth]{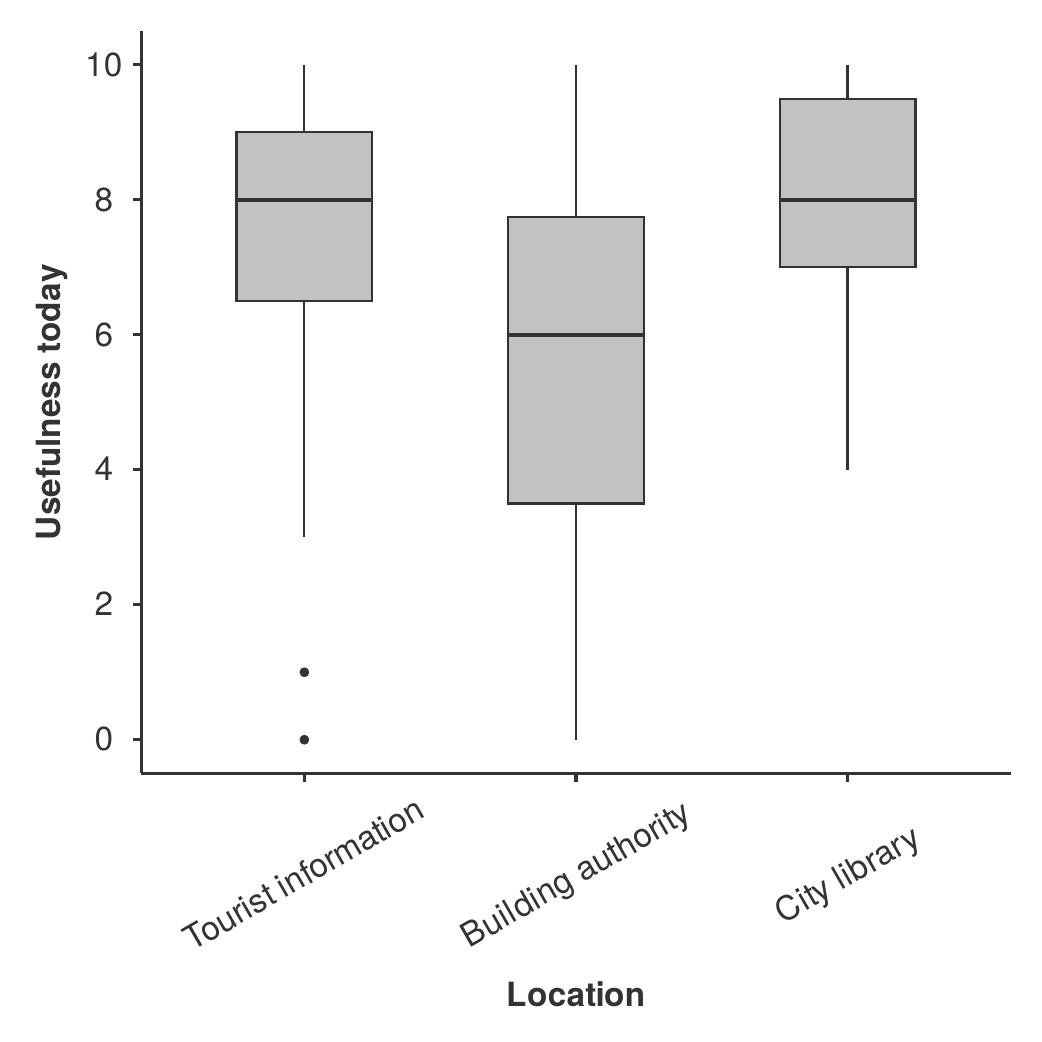}
    \caption{Box plot showing the distribution of the \textit{usefulness today} by location with indicated median values on a scale from zero to ten.}
    \label{usefulness_today_loc}
\end{minipage}
\end{figure*}

\section{Results} \label{sec:results}



\begin{table}
  \caption{Descriptive statistics of \textit{\acf{ITU}}, \textit{\acf{PU}}, \textit{\acf{PEOU}}, and \textit{usefulness today} by location.}
  \label{tab:table-by-location}
\includegraphics[scale=0.55]{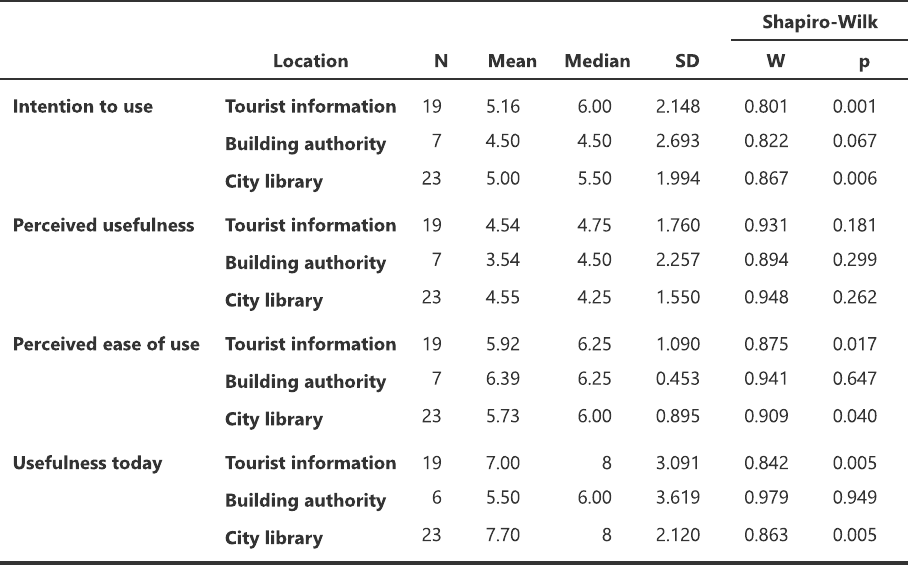}
\end{table}

All statistics were calculated using \textit{Jamovi} \cite{jamovi2024}.

\subsection{Previous knowledge of the robot}
Eighteen of the total of 49 interviewees knew about the availability of the robot prior to their visit (tourist information: five, building authority: six, city library: seven). Nine out of these 18 specifically came to the location to experience the robot live (tourist information: three, building authority: four, city library: two).

\subsection{Differences by location}
An analysis of the data categorized by location is presented in Figures~\ref{intention_to_use_loc}, \ref{perceived_usefulness_loc}, \ref{perceived_ease_of_use_loc}, and \ref{usefulness_today_loc}.
Table~\ref{tab:table-by-location} presents the descriptive statistics for each variable at each location, including the \textit{number of participants} (N), the \textit{mean}, the \textit{median}, and the \textit{standard deviation} (SD).

\begin{table}
  \caption{Results of the Kruskal–Wallis tests for \textit{\acf{ITU}}, \textit{\acf{PU}}, \textit{\acf{PEOU}}, and \textit{usefulness today}: for non of them $p < 0.05$ holds true}
  \label{kruskal_wallis}
\includegraphics[scale=0.85]{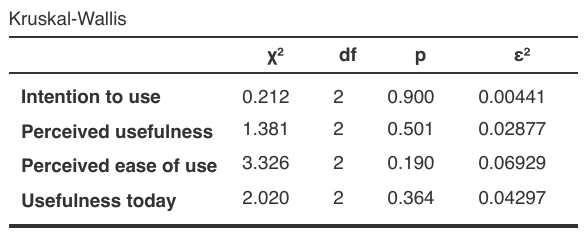}
\end{table}

Normality of the data was evaluated using the Shapiro–Wilk test for each variable and location. For several groups, the test indicated significant deviations from normal distribution ($p < 0.05$), including tourist information for ``\ac{ITU}'', ``\ac{PEOU}'', and ``How useful do you think would it be to use this robot here today?'' as well as city library for ``\ac{ITU}'', ``\ac{PEOU}'', and ``How useful do you think would it be to use this robot here today?''. As the assumption of normality was not fully met in all dimensions, a non-parametric Kruskal–Wallis test was applied to compare the locations. Table~\ref{kruskal_wallis} shows the obtained $p$-values. All of them exceeded the significance threshold ($\alpha = 0.05$), indicating that there are no statistically significant differences between the locations.

\begin{table}
  \caption{Results of the three emotion-related questions.}
  \label{emotions}
\includegraphics[scale=0.55]{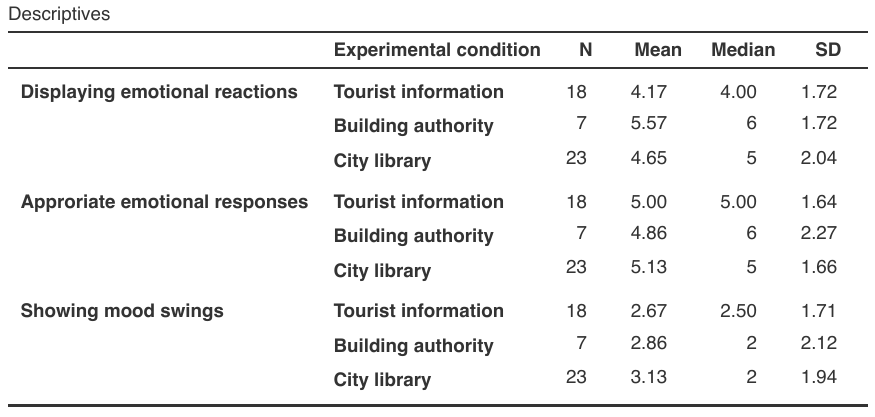}
\end{table}

\subsection{Perceived emotionality, interaction time, languages used, and suggestions}
The results for all three emotion-related questions do not differ significantly for the three locations, as shown in Table~\ref{emotions} with the mean, the median and the standard deviation for all items in all locations. The robot head was perceived as rather emotional (median value greater than four for the first two questions) and as not suffering from mood swings (median values less than four for the last question).

The self-reported interaction time varied from ``not at all'' to thirty minutes with an average interaction time of seven minutes. Most participants reported five minutes of previous interaction with the robot head, which is also the median of the distribution.

The multilingualism of the robot was used by many people during their interactions with the robot. The participants interacted with the robot in the following languages, with the frequency of use indicated in parentheses (some visitors tried out more than one language):
\textit{German (40), English (10), French (8), Spanish (3), Italian (3), Turkish (2), Arabic (1), Chinese (1), Croatian (1), Korean (1), Latvian (1), Persian (1), Polish (1), Portuguese (1), Russian (1), Swabian (a local German dialect) (1), }and\textit{ Vietnamese (1)}. 

Finally, the robot's long response time was frequently noted in the survey, with 10 out of 49 mentioning it. Others commented that the robot head appeared to be uncanny (five times) or funny/nice/innovative (seven times).

\section{Discussion and Limitations}\label{sec:limitations}
Our study has a couple of limitations that might impact its results. One possible explanation for the absence of statistically significant differences between the three locations is the relatively low number of participants, particularly at the building authority. This may have led to an insufficient, statistical power to detect differences across locations. 
Still, the descriptively lower \textit{Intention to use} and \textit{Usefulness today} at the building authority compared to both other locations are in line with a recent survey \cite{aymerich-franch_public_2026} that found robot deployments to be more acceptable in cultural and public leisure environments like libraries and museums than in government and administrative services.
Compared to a previous study with the full-body android robot Andrea employed in a public museum \cite{heisler2025conversationsandreavisitorsopinions}, the robot head was perceived as similarly useful in the tourist information and the city library, but less useful in the building authority location.
Together, this suggests that those findings from the above mentioned survey might generalize to real-world deployments of android robots although further investigation is needed for verification.
The fact that Andrea in the museum and Kim at the library and tourist information center are rated quite similarly raises questions about the effectiveness of the extensions made to the software. While the multilingualism was verbally appreciated by many users, it seems not to be reflected in the perceived usefulness of the robot. However, to draw clear conclusions in this regard other factors like appearance, location, and further extensions that had been implemented in the robot's software would need to be controlled for.

The uncontrolled interaction dynamics in such public environments with multiple people watching and interacting with the robot at the same time can be seen as another limitation. Sometimes visitors started to discuss their answers to the questionnaire items with other people around them. Therefore, the results of the present analysis should be complemented by similar studies in more controlled environments. Also, a long-term deployment would help assess sustained user acceptance over time. 

The choice of locations was influenced by practical considerations, e.g. their availability and willingness to accommodate an android robot. Similarly, the robot's clothing was influenced by suggestions of the hosting locations. In future studies different locations could be compared based on more theoretically grounded assumptions and stricter control of the robot head's appearance would be necessary, to eliminate possible uncontrolled influences of user perceptions through differing appearances of the robot head. 

\section{Conclusions}\label{sec:discussion_and_conclusion}

This exploratory study gathered the general public's opinion about a fully autonomous, very humanlike robot head that was set up for verbal interaction for several days each in three public spaces in Germany. Approximately one-third of the visitors (18 out of 49) knew about the availability of the robot in advance and half of them (9 out of 18) specifically came to experience the robot. This indicates a general interest in interacting with very human-like robots in public spaces. 

Overall, the robot head was evaluated at each location similarly positively on all three \ac{TAM2} subscales with median values greater than four on the standard seven point scale. 

It also became apparent that the system would benefit from a smoother interaction as was mentioned by every fifth visitor. The multilingualism of the robot was welcomed by everyone, who interacted with it, as is reflected in the wide variety of languages used.

Previous research suggests that in context-free interaction a more anthropomorphic design of a robot negatively impacts its likeability and perceived safety, but increases its attributed intelligence \cite{haring2016}. Therefore, investigating the technological acceptance of a less anthropomorphic, social robot in the same, public environments seems to be a logical next step.

In summary, the study underlines the potential of very human-like, robotic heads in service-oriented settings and contributes to the growing body of research on real-world \ac{HRI} in public places.

\balance

\bibliographystyle{IEEEtran}
\bibliography{references}


\end{document}